\documentclass{article}
\usepackage[utf8]{inputenc}
\usepackage{main}
\usepackage{microtype}
\usepackage{times}

\usepackage{amsmath,amsfonts,bm}

\def\eqref#1{equation~\ref{#1}}

\def\1{\bm{1}}

\DeclareMathAlphabet{\mathsfit}{\encodingdefault}{\sfdefault}{m}{sl}
\SetMathAlphabet{\mathsfit}{bold}{\encodingdefault}{\sfdefault}{bx}{n}

\usepackage{graphicx}
\usepackage{float}
\usepackage{wrapfig}
\usepackage{placeins}
\usepackage{url}
\usepackage{booktabs}
\usepackage{multirow}
\usepackage{colortbl}
\usepackage[most]{tcolorbox}
\usepackage{fancyhdr}
\usepackage{etoolbox}
\definecolor{darkblue}{rgb}{0,0.08,0.45}
\usepackage[colorlinks=true,linkcolor=darkblue,citecolor=darkblue,urlcolor=darkblue]{hyperref}
\floatstyle{ruled}
\newfloat{algorithm}{tbp}{loa}
\floatname{algorithm}{Algorithm}
\definecolor{routerrow}{RGB}{242,241,255}
\definecolor{bestrow}{RGB}{246,246,253}
\definecolor{deltagreen}{rgb}{0.0, 0.55, 0.0}
\definecolor{deltared}{rgb}{0.85, 0.0, 0.0}

\renewcommand{\headrulewidth}{0.7pt}
\newcounter{bibcount}
\makeatletter
\patchcmd{\@lbibitem}{\item[}{\item[\hfil\stepcounter{bibcount}{[\thebibcount]}}{}{}
\renewcommand\NAT@bibsetup[1]{%
  \setlength{\leftmargin}{\bibhang}%
  \setlength{\itemindent}{-\parindent}%
  \setlength{\itemsep}{\bibsep}%
  \setlength{\parsep}{\z@}}
\makeatother

\title{MOPD-Router: Rethinking Teacher Routing in \\ Multi-Teacher On-Policy Distillation}

\author{
\textbf{Tianze Xu}$^{1,4}$ \quad
\textbf{Yanzhao Zheng}$^{2}$ \quad
\textbf{Zhentao Zhang}$^{2}$ 
\textbf{Yuanqiang Yu}$^{2}$ \quad \\
\textbf{Chao Ma}$^{2}$ \quad 
\textbf{Jihuai Zhu}$^{2}$ \quad
\textbf{Lelun Wu}$^{2,5}$ \quad
\textbf{Lyumanshan Ye}$^{1,4}$ \quad
\textbf{Pengfei Liu}$^{1,3,4}$\thanks{Corresponding author} \\
\textbf{Baohua Dong}$^{2}$ \quad
\textbf{Hangcheng Zhu}$^{2}$ \quad
\textbf{Ruohui Huang}$^{2}$ \quad
\textbf{Gang Yu}$^{2}$\\[0.8em]
\textsuperscript{1}Shanghai Jiao Tong University\quad
\textsuperscript{2}Alibaba Group\quad
\textsuperscript{3}Shanghai Innovation Institute\quad \\
\textsuperscript{4}GAIR\quad
\textsuperscript{5}University of Science and Technology of China
}

\begin{document}

\maketitle

\fancypagestyle{firstpage}{%
    \fancyhf{}
    \lhead{\smash{\raisebox{-0.70cm}{\includegraphics[height=0.35cm]{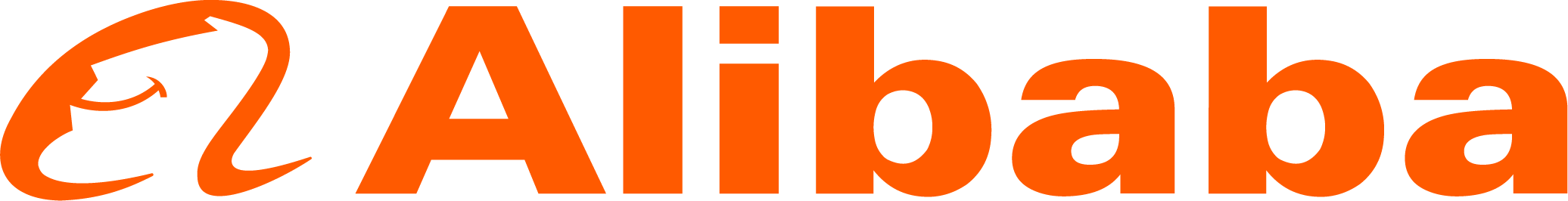}}}}
    \rhead{\smash{\raisebox{-0.87cm}{\includegraphics[height=0.7cm]{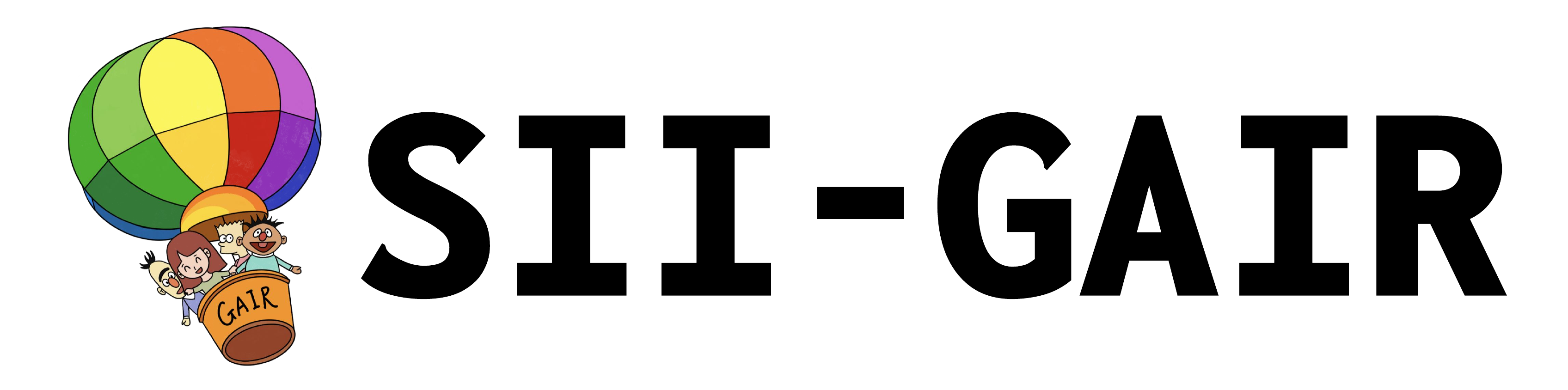}}}}
    \renewcommand{\headrulewidth}{0pt}
}
\thispagestyle{firstpage}

\begin{abstract}
Multi-teacher on-policy distillation (MOPD) integrates specialized capabilities into a single student, but existing practice typically hard-routes each prompt to a domain-matched teacher for the entire rollout. This dependence on prompt-level domain labels restricts using unlabeled training mixtures and leaves complementary signals from other teachers unused. We introduce \textbf{MOPD-Router}, a framework that routes supervision over the full teacher pool at each token, without domain labels or training a separate routing model. Its plug-in interface supports different metrics for selecting and weighting teacher-specific OPD signals. Within this interface, we propose \textbf{ExpertAlign}, which scores each teacher by whether its correction to the student at the current token expresses the specialization that teacher acquired during post-training, and compare it against two reference metrics built on teacher confidence (Entropy) and teacher--student discrepancy (Novelty). Experiments on unlabeled and domain-labeled training mixtures under strong-to-weak and same-size distillation scenarios show that ExpertAlign achieves the strongest overall performance in all four settings. On unlabeled data, it improves the overall score by 5.88 (+12.3\%) points over Mean aggregation; on domain-labeled data, it outperforms standard MOPD by 3.95 (+7.8\%) points without using available domain labels. These results demonstrate token-level routing can exploit cross-domain complementary supervision, and reduce exclusive reliance on prompt-level domain assignment.\footnote{Code is available at: \url{https://github.com/TURLEing/MOPD-Router}.}
\end{abstract}

\begin{figure}[H]
    \centering
    \includegraphics[width=0.89\linewidth]{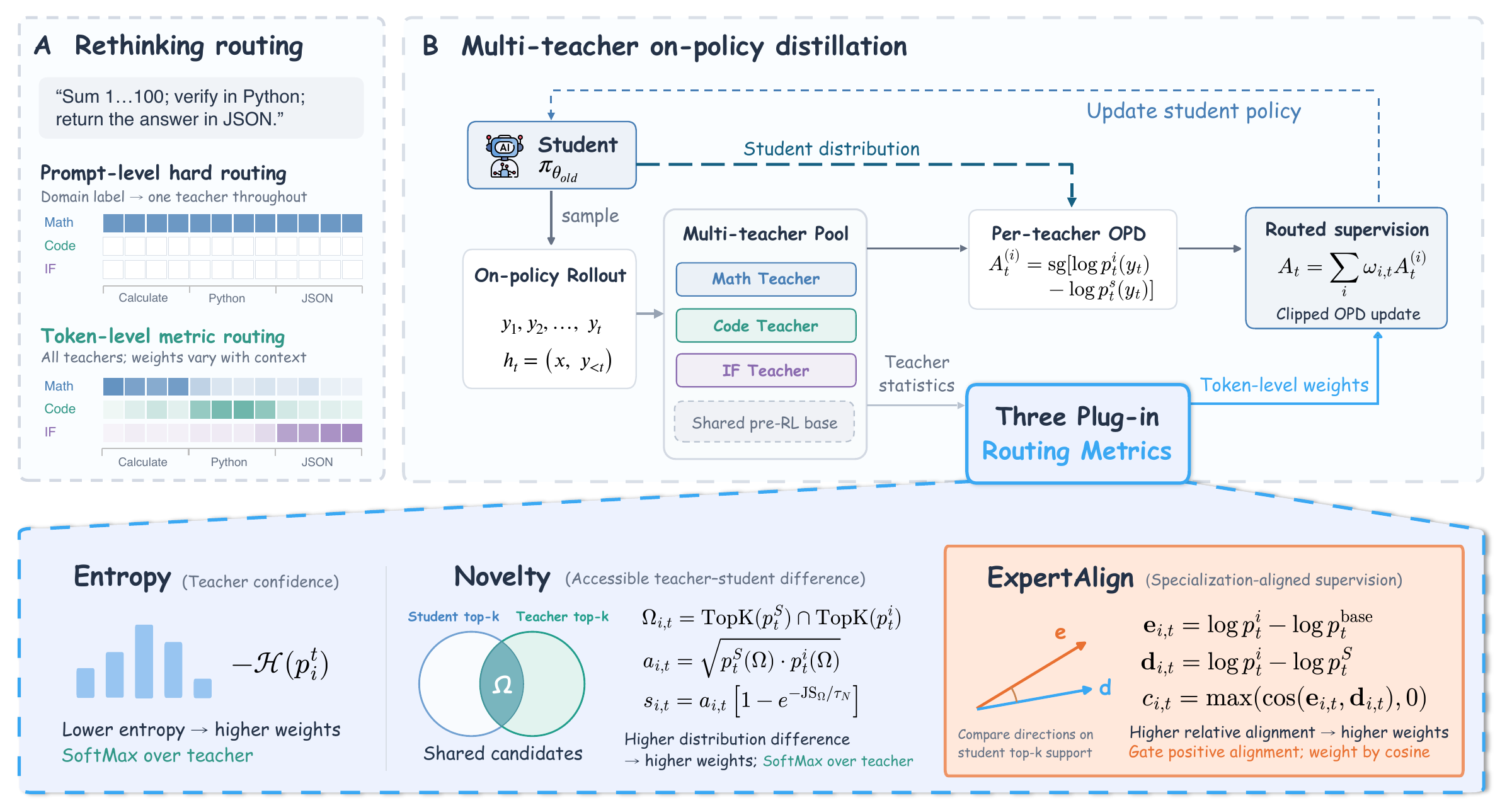}
    \small
    \caption{\textbf{Overview of MOPD-Router.} (a) Unlike prompt-level hard routing, token-level routing dynamically combines supervision from the full teacher pool. (b) Teachers evaluate the student's on-policy rollout, while a plug-in metric weights their token-level OPD signals. (c) Three routing metrics: teacher confidence (Entropy), teacher--student difference (Novelty), and specialization--direction alignment (ExpertAlign).}
    \label{fig:mopd_router_overview}
\end{figure}

\section{Introduction}

Reinforcement learning (RL) has emerged as an important approach to improving capabilities of large language model (LLM) ~\citep{ouyang2022instructgpt,schulman2017ppo,shao2024deepseekmath}. Different task domains often use distinct RL pipelines: mathematical reasoning can exploit verifiable rewards, software engineering can train in executable environments, and instruction following can use rule- or rubric-based feedback~\citep{ma2026mopd}. These independent RL pipelines, often starting from a shared base model, produce specialist models with distinct capabilities. Yet the ultimate goal is a single model that performs well across these domains, making the integration of multiple RL-specialized capabilities central to LLM post-training~\citep{ma2026mopd,gao2026openmopd}.

Nowadays, Multi-teacher On-Policy Distillation (MOPD) is an appealing paradigm for integrating specialized capabilities into one policy model~\citep{ma2026mopd,gao2026openmopd}. MOPD builds on On-Policy Distillation (OPD), where the student generates trajectories under its current policy and teachers provide dense token-level supervision on them~\citep{agarwal2024opd,yang2026gopd,li2026rethinkingopd}. Adopting MOPD for capability integration is now a consensus among leading LLM vendors~\citep{xiao2026mimov2flash,zeng2026glm5,xu2026deepseekv4,kimi2026k3}. However, existing MOPD protocol typically uses a coarse-grained teacher assignment: it selects the single teacher for each prompt corresponding to the domain label, then the teacher supervises the entire rollout through OPD~\citep{ma2026mopd,gao2026openmopd}. We refer to this one-teacher-per-prompt protocol as \emph{domain-label hard routing}. It can identify the primary domain expert and provide stable MOPD signals, but relies on prompt-level label for teacher assignment throughout the trajectory.

This protocol has two limitations. First, real-world mixed datasets of SFT and chat often lack reliable domain labels. A prompt may combine general conversation, instruction following, mathematical reasoning, and code execution, leaving standard MOPD no predefined rule for assigning one domain teacher. Second, a domain label identifies the primary teacher but ignores the potential value of other teachers at individual tokens. Taking mathematics prompts as an example, a code teacher may help with programmatic reasoning, while an instruction-following teacher may help satisfy format or output constraints. The usefulness of these signals can also vary across generation stages and token positions. Routing the entire rollout to a fixed domain teacher therefore leaves potential cross-domain complementary supervision unused.

However, exploiting this potential cross-domain complementarity requires token- and context-aware routing. This raises a key question: \textbf{how can a routing metric identify and select relevant teachers and assign appropriate weights to their signals at each token?}

Thus, we introduce \textbf{MOPD-Router}, a token-level teacher-routing framework for MOPD that works without domain labels or training a separate routing model, providing an interface for studying routing metrics within a same MOPD pipeline.We instantiate three metrics from different information sources. \emph{Entropy} routes by a teacher's next-token entropy, and \emph{Novelty} by the accessible teacher--student distributional difference. Beyond that, we introduce \textbf{ExpertAlign}, which routes each token by the alignment between the teacher's acquired specialization and its teaching supervision. Figure~\ref{fig:mopd_router_overview} gives an overview of the framework and metrics.

We evaluate MOPD-Router on two training datasets under both strong-to-weak and same-size OPD scenarios. On the unlabeled mixed dataset, it enables multi-teacher OPD without predefined domain assignment, with effective routing metrics outperforming naive mean aggregation. On the domain-labeled MOPD dataset, token-level metric routing matches or exceeds domain-label hard without using available labels, suggesting that effective routing need not rely exclusively on prompt-level domain labels. Among these routing metrics, our proposed \emph{ExpertAlign} achieves the strongest overall performance across the evaluated data and model-scale settings.

Our contributions are as follows:
\begin{itemize}
    \item We introduce \textbf{MOPD-Router}, a plug-in framework that selects and weights supervision signals from multi-teacher pool at every token, without using domain labels or training a separate routing model.
    \item We propose \textbf{ExpertAlign}, a specialization-aware routing metric that aligns a teacher's acquired expertise direction with the teaching direction it would apply to the current student. Across unlabeled and domain-labeled data under two distillation scenarios, ExpertAlign achieves the best overall performance in all four settings.
    \item Our analyses show that cross-domain teacher signals provide complementary supervision across capabilities. We open-source the MOPD-Router framework to support further research on token-level teacher routing.
\end{itemize}

\section{Related Work}

\subsection{On-Policy Distillation}

On-policy distillation (OPD) combines the distributional consistency of learning from student-generated trajectories with dense token-level teacher supervision~\citep{agarwal2024opd}. Recent OPD research has three directions~\citep{yu2026dopd}: (a) \emph{Strong-to-weak distillation} transfers capabilities from a stronger teacher to a weaker student and develops more stable or effective variants of the standard OPD objective~\citep{agarwal2024opd,yang2026gopd}. (b) \emph{Self-distillation} uses the same model as teacher and student under different context or feedback conditions~\citep{hubotter2026selfdistillation,zhao2026opsd}. (c) \emph{Adaptive distillation} addresses non-uniform OPD supervision by modulating the training objective or supervised tokens using different signals~\citep{jin2026eopd,wang2026teachability,xu2026tip,li2026rethinkingopd,heo2026opd2}. These analyses and objectives inspire our routing designs, but primarily study modifying supervision within single-teacher OPD rather than allocating it among multiple specialized teachers.

\subsection{Multi-Teacher Capability Integration}

Prior approaches integrate capabilities through joint training, sequential training, or weight-space composition~\citep{ma2026mopd}. Mix-RL pools prompts from multiple domains and jointly optimizes their domain-specific rewards~\citep{yang2025qwen3}. Cascade RL applies domain-specific RL stages sequentially to a single model~\citep{wang2026nemotroncascade}. Parameter merging instead combines independently specialized checkpoints directly in weight space~\citep{ilharco2023taskarithmetic}.

Multi-teacher OPD (MOPD) integrates independently trained domain teachers in policy space through dense supervision on student-generated trajectories~\citep{ma2026mopd}. It has been adopted in several frontier-model post-training pipelines, including MiMo-V2-Flash, GLM-5, DeepSeek-V4, Kimi K3, and Nemotron-Cascade 2~\citep{xiao2026mimov2flash,zeng2026glm5,xu2026deepseekv4,kimi2026k3,yang2026nemotroncascade2}. Open-MOPD and D$^3$-MOPD further increase the fraction of domain-teacher gains recovered by the student through optimization-budget balancing and adaptive domain scheduling~\citep{gao2026openmopd,sun2026d3mopd}. H-OPD introduces confidence-based token-level arbitration between vision-language and text-only teachers for multimodal reasoning~\citep{yin2026hopd}. However, most existing MOPD implementations still rely on prompt-level hard routing based on domain labels, leaving label-free routing metric design underexplored.

\section{MOPD-Router}

MOPD-Router replaces fixed prompt-to-teacher assignment with token-level weighting over the full teacher set. We formalize sampled-token MOPD as routing over per-teacher OPD advantages, then instantiate three independent metrics drawing on the teacher distribution, the teacher--student relationship, and teacher specialization relative to the shared pre-RL base model.

\subsection{Multi-Teacher On-Policy Distillation}

Let $x\sim\mathcal{D}$ be a prompt, $\pi_{\theta}$ the student policy, and $\{\pi_{\phi_i}\}_{i=1}^{M}$ the expert teachers. The rollout policy $\pi_{\theta_{\mathrm{old}}}$ samples a response $y=(y_1,\ldots,y_T)\sim\pi_{\theta_{\mathrm{old}}}(\cdot\mid x)$. At position $t$, a teacher evaluates the student-visited context $h_t=(x,y_{<t})$. Following sampled-token OPD implementations~\citep{li2026rethinkingopd}, we define per-teacher and routed advantages as
\begin{equation}
    A^{(i)}_t
    =\operatorname{sg}\!\left[
        \log \pi_{\phi_i}(y_t\mid h_t)
        -\log \pi_{\theta_{\mathrm{old}}}(y_t\mid h_t)
    \right],
    \qquad
    A_t=\sum_{i=1}^{M}w_{i,t}A^{(i)}_t.
    \label{eq:mopd_advantage}
\end{equation}
Here $\operatorname{sg}[\cdot]$ denotes stop-gradient and $A_t$ aggregates the per-teacher sampled-token signals $A^{(i)}_t$ with non-negative weights $\{w_{i,t}\}_{i=1}^{M}$. As in RL-style OPD implementations, $A_t$ directly occupies the advantage slot without a learned critic~\citep{heo2026opd2,gao2026openmopd}. Given the importance ratio $\rho_t(\theta)=\pi_{\theta}(y_t\mid h_t)/\pi_{\theta_{\mathrm{old}}}(y_t\mid h_t)$, the student maximizes~\citep{schulman2017ppo}
\begin{equation}
   \mathcal{J}_{\mathrm{MOPD}}(\theta)
   =\mathbb{E}\!\left[
       \frac{1}{T}\sum_{t=1}^{T}
       \min\!\left(
           \rho_t(\theta)A_t,
           \operatorname{clip}(\rho_t(\theta),1-\epsilon,1+\epsilon)A_t
       \right)
   \right].
   \label{eq:mopd_objective}
\end{equation}
Standard MOPD assigns each prompt a domain label $d(x)$ and uses the corresponding teacher throughout the response~\citep{ma2026mopd,gao2026openmopd}. This \emph{domain-label hard routing} uses
\begin{equation}
    w^{\mathrm{domain}}_{i,t}=\mathbb{I}[i=d(x)],
    \label{eq:domain_hard_routing}
\end{equation}
reusing the same one-hot weights at every position. When no domain assignments are available, a direct alternative is \emph{Mean} aggregation, with $w^{\mathrm{mean}}_{i,t}=1/M$. Domain-label hard routing is reliable but coarse, whereas Mean aggregation is label-free but treats all teacher signals as equally informative.

\subsection{Token-Level Metric Routing}

MOPD-Router replaces these fixed aggregation rules by evaluating a plug-in metric $\mathcal{M}$ for every teacher at each token position of the student rollout:
\begin{equation}
    s_{i,t}=\mathcal{M}\!\left(\pi_{\phi_i},\pi_{\theta_{\mathrm{old}}},\pi_{\theta_{\mathrm{base}}};h_t\right),
    \label{eq:metric_router}
\end{equation}
where $\pi_{\theta_{\mathrm{base}}}$ denotes the teachers' shared base model before post-training. At each token, metric routing uses scores to identify suitable teachers and assign their relative contributions to $A_t$, yielding context-dependent selection and weighting over the full teacher pool without domain labels or training a separate routing model. Algorithm~\ref{alg:mopd_router} summarizes the training loop.

\subsection{Routing Metrics}

Let $p^S_t(v)=\pi_{\theta_{\mathrm{old}}}(v\mid h_t)$ denote the rollout student's next-token distribution, $p^i_t(v)=\pi_{\phi_i}(v\mid h_t)$ teacher $i$'s distribution, and $p^{\mathrm{base}}_t(v)=\pi_{\theta_{\mathrm{base}}}(v\mid h_t)$ the shared base model's distribution. We study three routing metrics for estimating token-level teacher relevance. Because the Entropy and Novelty scores occupy narrow metric-specific scales, their softmax weights use sub-unit routing temperatures to avoid near-uniform weights and sharpen distinctions among teachers.

\paragraph{Entropy.}
A natural but naive hypothesis is that a more confident teacher should receive greater routing weight. Entropy measures how diffuse a teacher's next-token distribution is: lower entropy concentrates probability mass on fewer candidate tokens and therefore indicates a sharper, more confident prediction. Entropy, representing the confidence-arbitration method~\citep{yin2026hopd,prabhudesai2025rent}, uses negative predictive entropy as the routing score
\begin{equation}
    s^{\mathrm{Ent}}_{i,t}
    =-H(p^i_t)
    =\sum_{v\in\mathcal{V}}p^i_t(v)\log p^i_t(v),
    \qquad
    w^{\mathrm{Ent}}_{i,t}
    =\frac{\exp(s^{\mathrm{Ent}}_{i,t}/\tau_R^{\mathrm{Ent}})}
    {\sum_{j=1}^{M}\exp(s^{\mathrm{Ent}}_{j,t}/\tau_R^{\mathrm{Ent}})},
    \label{eq:entropy_score}
\end{equation}
where $\tau_R^{\mathrm{Ent}}=0.1$ is the routing Softmax temperature. This teacher-only metric is computationally straightforward, but cannot determine whether a lower-entropy teacher usefully corrects the current student, and intrinsic entropy-scale differences can systematically favor particular teachers.

\paragraph{Novelty.}
Prior work indicates that effective OPD updates concentrate on high-probability candidates shared by teacher and student, whereas corrections outside the student's local support can be difficult to absorb~\citep{li2026rethinkingopd,wang2026teachability}. \emph{Novelty} therefore assigns a high score when teacher differs substantially from the student within their shared high-probability candidate set and both assign it sufficient probability mass

Let $S^S_t=\operatorname{TopK}(p^S_t,k)$ and $S^i_t=\operatorname{TopK}(p^i_t,k)$ denote their sets of $k$ most probable next-token candidates, and let $\Omega_{i,t}=S^S_t\cap S^i_t$ be their shared candidate set. For a distribution $p$, we define $p(\Omega)=\sum_{v\in\Omega}p(v)$ and $p|_{\Omega}$ for its restriction to $\Omega$, renormalized to unit mass. We also define the \emph{accessibility} and \emph{novelty} components as
\begin{equation}
    a_{i,t}=\sqrt{p^S_t(\Omega_{i,t})p^i_t(\Omega_{i,t})},
    \qquad
    N_{i,t}=1-\exp\!\left(-\operatorname{JS}\!\left(p^S_t|_{\Omega_{i,t}},p^i_t|_{\Omega_{i,t}}\right)/{\tau_N}\right).
    \label{eq:novelty_components}
\end{equation}
The \emph{accessibility} term $a_{i,t}$ measures both models' probability mass on shared candidate tokens, while the \emph{novelty} term $N_{i,t}$ measures their distributional discrepancy within the shared candidate set. The temperature $\tau_N=0.1$ controls how quickly this discrepancy saturates. We combine both components and instantiate Novelty routing as
\begin{equation}
    s^{\mathrm{Nov}}_{i,t}=a_{i,t}N_{i,t},
    \qquad
    w^{\mathrm{Nov}}_{i,t}=\frac{\exp(s^{\mathrm{Nov}}_{i,t}/\tau_R^{\mathrm{Nov}})}{\sum_{j=1}^{M}\exp(s^{\mathrm{Nov}}_{j,t}/\tau_R^{\mathrm{Nov}})}.
    \label{eq:novelty_score}
\end{equation}
We use $\tau_R^{\mathrm{Nov}}=0.1$ and set $s^{\mathrm{Nov}}_{i,t}=0$ when $\Omega_{i,t}$ is empty.

\paragraph{ExpertAlign.}
Entropy and Novelty score a teacher's current output, but do not indicate whether it reflects the teacher’s expertise. Prior work~\citep{heo2026opd2,feng2026directopd} shows that the log-policy difference between a teacher and its base model captures the policy shift induced by post-training. We therefore propose \emph{ExpertAlign}, a specialization-aware token-level routing metric that assesses whether a teacher would move the current student in the direction of its post-training expertise. Building on this observation, we first use the base-relative shift to define teacher $i$'s \emph{expertise vector} on the student's top-$k$ support $S^S_t$:
\begin{equation}
    \mathbf{e}_{i,t}
    =\left[
        \log p^i_t(v)-\log p^{\mathrm{base}}_t(v)
    \right]_{v\in S^S_t},
    \label{eq:expertise_vector}
\end{equation}
which represents the direction acquired by teacher $i$ relative to the shared pre-RL base and is an intrinsic property of that teacher. Against it, we further introduce the teacher--student \emph{teaching vector}, a student-dependent direction defined as
\begin{equation}
    \mathbf{d}_{i,t}
    =\left[
        \log p^i_t(v)-\log p^S_t(v)
    \right]_{v\in S^S_t},
    \label{eq:teaching_vector}
\end{equation}
to capture how teacher $i$ would reshape the student's log probabilities over its local top-$k$ support. \textbf{ExpertAlign considers teacher $i$ relevant only when its teaching direction $\mathbf{d}_{i,t}$ aligns with the specialization $\mathbf{e}_{i,t}$}, indicating that the correction reflects capabilities acquired during teacher post-training rather than generic disagreement with the student. We thus quantify this specialization--teaching alignment using positive cosine similarity:
\begin{equation}
    c_{i,t}
    =\max\!\left(
        \frac{\langle\mathbf{e}_{i,t},\mathbf{d}_{i,t}\rangle}
        {\lVert\mathbf{e}_{i,t}\rVert_2\lVert\mathbf{d}_{i,t}\rVert_2},
        0
    \right),
    \qquad
    \mathcal{E}_t=\{i:\langle\mathbf{e}_{i,t},\mathbf{d}_{i,t}\rangle>\delta\},
    \label{eq:expert_align_alignment}
\end{equation}
where a zero-norm vector receives $c_{i,t}=0$, and the numerical alignment margin is $\delta=10^{-6}$. \emph{ExpertAlign} only retains teachers whose teaching signal positively aligns with their acquired specialization. We consider two weighting rules over the retained set:
\begin{equation}
    w^{\mathrm{uniform}}_{i,t}
    =\frac{\mathbb{I}[i\in\mathcal{E}_t]}{|\mathcal{E}_t|},
    \qquad
    w^{\mathrm{cosine}}_{i,t}
    =\frac{\mathbb{I}[i\in\mathcal{E}_t]c_{i,t}}
    {\sum_{j\in\mathcal{E}_t}c_{j,t}}.
    \label{eq:expert_align_weights}
\end{equation}
If $\mathcal{E}_t$ is empty, all weights are zero and OPD supervision at position $t$ is skipped. We select $k=16$ and report its sensitivity in Appendix~\ref{app:expert_align_topk_sensitivity}. Appendix~\ref{app:expert_align_skip} further reports the retained-teacher distributions and their evolution across all four settings. 

\begin{algorithm}[t]
\caption{MOPD-Router training loop}
\label{alg:mopd_router}
\small
\textbf{Require:} Prompt sampler $\mathcal{D}$, student $\pi_\theta$, teachers $\{\pi_{\phi_i}\}_{i=1}^{M}$, shared teacher base $\pi_{\theta_{\mathrm{base}}}$, routing metric $\mathcal{M}$, inner-step count $K$\\
\textbf{Ensure:} Updated student $\pi_\theta$\\[-0.25em]
\begin{tabular}{@{}r@{\hspace{0.7em}}p{0.92\linewidth}@{}}
1: & \textbf{for} each training step $q$ \textbf{do} \\
2: & \quad Set $\pi_{\theta_{\mathrm{old}}}\leftarrow\pi_\theta$ and sample a rollout batch $\mathcal{R}_q$ from $\mathcal{D}$ and $\pi_{\theta_{\mathrm{old}}}$. \\
3: & \quad Evaluate the student, all teachers, and the shared base on the student-visited contexts to obtain the quantities required by $\mathcal{M}$. \\
4: & \quad Compute each teacher's sampled-token advantage $A_t^{(i)}$ using Equation~\ref{eq:mopd_advantage}. \\
5: & \quad Apply $\mathcal{M}$ to obtain non-negative token-level teacher weights $\{w_{i,t}\}_{i=1}^{M}$ over the full teacher pool. \\
6: & \quad Aggregate the routed advantage $A_t=\sum_{i=1}^{M}w_{i,t}A_t^{(i)}$. \\
7: & \quad \textbf{for} each inner minibatch $\mathcal{B}_{q,k}$, $k=0,\ldots,K-1$ \textbf{do} \\
8: & \qquad Recompute current-student log-probabilities and the importance ratios $\rho_t(\theta)$. \\
9: & \qquad Update $\theta$ by maximizing the clipped MOPD objective in Equation~\ref{eq:mopd_objective}. \\
10: & \quad \textbf{end for} \\
11: & \textbf{end for}
\end{tabular}
\end{algorithm}

\section{Experiments}

We evaluate MOPD-Router across two training-data regimes and two student–teacher distillation scenarios. We first test metric-based routing on a mixed dataset without teacher-assignment labels, then compare it with standard domain-label hard routing on a conventional labeled MOPD dataset. We further analyze teacher selection, metric design, and cross-domain supervision.

\subsection{Experimental Setup}

\paragraph{Training data.}
Our first training set is a 60K unlabeled mixture of 30K general chatting prompts sampled from Nemotron-v3 Dataset~\citep{nvidia2026nemotronsft} and 30K mathematics prompts from NuminaMath-TIR~\citep{aimo2024numinamath} requiring executable code for problem solving. It lacks teacher-assignment labels and thus cannot directly use domain-label hard routing. Our second training set is a domain-labeled post-training dataset comprising 25K mathematics prompts from DeepMath-103K~\citep{he2025deepmath}, 25K code prompts from Eurus-RL-Code~\citep{cui2025prime}, and 16K instruction-following prompts from HiR-16K~\citep{zhang2025hir}. It represents the conventional MOPD setting, where each prompt is assigned to its corresponding teacher; MOPD-Router ignores these labels and routes over all teachers using only its token-level metric.

\begin{table}[t]
    \caption{Results (\%) on the unlabeled mixed training set. Bold denotes the best distilled result within each student block; shaded rows identify MOPD-Router methods. The final row reports the same-size ExpertAlign result minus Standard MOPD.}
    \label{tab:unlabeled_main}

    \centering
    \setlength{\tabcolsep}{2.6pt}
    \renewcommand{\arraystretch}{1.0}
    \resizebox{\textwidth}{!}{%
    \begin{tabular}{l*{10}{c}}
        \toprule
        & \multicolumn{4}{c}{Math} & \multicolumn{3}{c}{Code} & \multicolumn{2}{c}{IF} & \multicolumn{1}{c}{Overall} \\
        \cmidrule(lr){2-5}\cmidrule(lr){6-8}\cmidrule(lr){9-10}\cmidrule(lr){11-11}
        Method & \small{AIME24} & \small{AIME25} & \small{\shortstack{HMMT\\Feb.}} & \small{\shortstack{HMMT\\Nov.}} & \small{\shortstack{Human\\Eval+}} & \small{MBPP+} & \small{LCB-v6} & \small{IFEval} & \small{IFBench} & Avg. \\
        \midrule
        \multicolumn{11}{c}{\textit{Baseline Models}} \\
        \midrule
        Qwen3-1.7B & 12.50 & 11.67 & 5.83 & 4.58 & 60.40 & 52.40 & 12.00 & 67.65 & 16.67 & 27.08 \\
        Qwen3-4B & 23.33 & 10.00 & 10.00 & 7.08 & 82.30 & 67.20 & 18.29 & 81.88 & 24.14 & 36.02 \\
        RL Teacher  & 58.33 & 52.08 & 28.75 & 35.00 & 84.14 & 72.20 & 24.57 & 82.80 & 31.29 & 52.13 \\
        \midrule
        \multicolumn{11}{c}{\textit{Student: Qwen3-1.7B-Non-Thinking}} \\
        \midrule
        Standard MOPD & 32.08 & 25.41 & 16.67 & 15.41 & 70.12 & 57.41 & 19.42 & 69.32 & 21.09 & 36.33 \\
        Open-MOPD & 24.58 & 25.42 & 15.00 & 12.08 & 71.95 & 56.35 & 13.14 & 66.91 & 20.75 & 34.02 \\
        Mean Aggregation & 27.92 & 26.25 & 15.42 & 11.67 & 70.73 & 53.17 & 16.57 & 70.98 & 17.68 & 34.49 \\
        \addlinespace[0.25\baselineskip]
        \rowcolor{routerrow}MOPD Router & & & & & & & & & & \\
        \rowcolor{routerrow}~~w/ Entropy & 30.00 & 26.67 & 15.42 & 13.33 & 68.90 & \textbf{59.52} & 15.43 & 64.51 & 18.03 & 34.65 \\
        \rowcolor{routerrow}~~w/ Novelty & 32.92 & 27.92 & \textbf{17.08} & 13.75 & 70.73 & 57.67 & \textbf{20.00} & \textbf{71.90} & 18.37 & 36.70 \\
        \rowcolor{routerrow}~~\textbf{w/ ExpertAlign} & \textbf{38.33} & \textbf{32.08} & 16.67 & \textbf{17.08} & \textbf{72.56} & 58.99 & 19.43 & 70.61 & \textbf{21.43} & \textbf{38.58} \\
        \midrule
        \multicolumn{11}{c}{\textit{Student: Qwen3-4B-Non-Thinking}} \\
        \midrule
        Standard MOPD & 50.00 & 45.83 & 25.41 & 33.75 & 84.15 & 69.58 & 28.00 & 77.26 & 24.83 & 48.76 \\
        Open-MOPD & 51.25 & 45.41 & 22.92 & 32.92 & 86.59 & 71.96 & 24.57 & 81.80 & 27.27 & 49.41 \\
        Mean Aggregation & 51.25 & 45.83 & 22.50 & 29.58 & 82.93 & 69.84 & 26.28 & 80.22 & 22.45 & 47.88 \\
        \addlinespace[0.25\baselineskip]
        \rowcolor{routerrow}MOPD Router & & & & & & & & & & \\
        \rowcolor{routerrow}~~w/ Entropy & 51.67 & 48.33 & 22.50 & 31.25 & 81.71 & 71.96 & 26.86 & 78.19 & 20.07 & 48.06 \\
        \rowcolor{routerrow}~~w/ Novelty & 56.25 & 44.17 & 26.66 & 35.83 & 84.76 & 70.63 & \textbf{29.71} & 79.85 & 24.83 & 50.30 \\
        \rowcolor{routerrow}~~\textbf{w/ ExpertAlign} & \textbf{59.17} & \textbf{52.08} & \textbf{33.75} & \textbf{40.42} & \textbf{87.20} & \textbf{73.02} & 28.57 & \textbf{82.07} & \textbf{27.55} & \textbf{53.76} \\
        \midrule
        \rowcolor{bestrow}$\Delta$ vs Standard MOPD & \textcolor{deltagreen}{+9.17} & \textcolor{deltagreen}{+6.25} & \textcolor{deltagreen}{+8.34} & \textcolor{deltagreen}{+6.67} & \textcolor{deltagreen}{+3.05} & \textcolor{deltagreen}{+3.44} & \textcolor{deltagreen}{+0.57} & \textcolor{deltagreen}{+4.81} & \textcolor{deltagreen}{+2.72} & \textcolor{deltagreen}{\textbf{+5.00}} \\
        \bottomrule
    \end{tabular}
    }
\end{table}

\paragraph{Models and teachers.}
We use the non-thinking variants of Qwen3 series~\citep{yang2025qwen3}. The teacher pool comprises three Qwen3-4B-Non-Thinking models obtained through domain-specific RL post-training. The mathematics and code teachers are the publicly released RL checkpoints from ExOPD~\citep{yang2026gopd}; we train the instruction-following teacher on HIR-16K~\citep{zhang2025hir}, detailed in Appendix~\ref{app:implementation_details}. All teachers share Qwen3-4B-Non-Thinking as their pre-RL base model. We consider two distillation settings: \emph{strong-to-weak}, with a Qwen3-1.7B student and Qwen3-4B teachers, and \emph{same-size}, with a Qwen3-4B student and Qwen3-4B teachers.

\paragraph{Baselines.}
In both distillation scenarios, we compare against three baselines: \emph{Standard MOPD}, \emph{Open-MOPD}~\citep{gao2026openmopd}, and \emph{Mean Aggregation}, which uniformly averages the three sampled-token teacher signals. On the unlabeled mixture, DeepSeek-V4-Flash~\citep{xu2026deepseekv4} labels every SFT example as math, code, or instruction following from its prompt and reference response. Standard MOPD uses these labels for teacher assignment, while Open-MOPD uses the same assignments and balances the optimization budget across domains. Appendix~\ref{app:ai_domain_labels} reports the prompt and label distribution. On the domain-labeled dataset, Standard MOPD and Open-MOPD use the provided domain labels. We compare these baselines with Entropy, Novelty, and ExpertAlign with cosine weighting. We also report the initial student and a \emph{Teacher reference} evaluating each teacher only on its respective domain.

\paragraph{Training and evaluation.}
Within each setting, all routing methods use the same sampled-token MOPD objective, teacher pool, rollout configuration, and optimization budget. We evaluate math on AIME 2024~\citep{aimo2024aime}, AIME 2025~\citep{opencompass2025aime}, HMMT 2025 Feb~\citep{balunovic2025matharena}, and HMMT 2025 Nov~\citep{balunovic2025matharena}; code on HumanEval+, MBPP+~\citep{liu2023evalplus} and LiveCodeBench v6~\citep{jain2024livecodebench}; and instruction following on IFEval~\citep{zhou2023ifeval} and IFBench~\citep{pyatkin2025ifbench}. We sample 8 responses per problem and report Avg@8 accuracy. Code tasks use Pass@1; instruction-following tasks use five independent completions per instance and report prompt-level accuracy. \emph{Overall Avg.} macro-averages the nine benchmark metrics. Appendix~\ref{app:implementation_details} lists all training and evaluation hyperparameters.

\subsection{Main Results}

\begin{table}[t]
    \caption{Results (\%) on the domain-labeled MOPD training set. Formatting follows Table~\ref{tab:unlabeled_main}.}
    \label{tab:labeled_main}
    \centering
    \setlength{\tabcolsep}{2.6pt}
    \renewcommand{\arraystretch}{1.0}
    \resizebox{\textwidth}{!}{%
    \begin{tabular}{l*{10}{c}}
        \toprule
        & \multicolumn{4}{c}{Math} & \multicolumn{3}{c}{Code} & \multicolumn{2}{c}{IF} & \multicolumn{1}{c}{Overall} \\
        \cmidrule(lr){2-5}\cmidrule(lr){6-8}\cmidrule(lr){9-10}\cmidrule(lr){11-11}
        Method & \small{AIME24} & \small{AIME25} & \small{\shortstack{HMMT\\Feb.}} & \small{\shortstack{HMMT\\Nov.}} & \small{\shortstack{Human\\Eval+}} & \small{MBPP+} & \small{LCB-v6} & \small{IFEval} & \small{IFBench} & Avg. \\
        \midrule
        \multicolumn{11}{c}{\textit{Baseline Models}} \\
        \midrule
        Qwen3-1.7B & 12.50 & 11.67 & 5.83 & 4.58 & 60.40 & 52.40 & 12.00 & 67.65 & 16.67 & 27.08 \\
        Qwen3-4B & 23.33 & 10.00 & 10.00 & 7.08 & 82.30 & 67.20 & 18.29 & 81.88 & 24.14 & 36.02 \\
        RL Teacher  & 58.33 & 52.08 & 28.75 & 35.00 & 84.14 & 72.20 & 24.57 & 82.80 & 31.29 & 52.13 \\
        \midrule
        \multicolumn{11}{c}{\textit{Student: Qwen3-1.7B-Non-Thinking}} \\
        \midrule
        Standard MOPD & 33.33 & 34.17 & 15.83 & 16.67 & 69.51 & 61.11 & 19.43 & 68.21 & 19.39 & 37.52 \\
        Open-MOPD & 35.42 & \textbf{35.00} & 17.91 & 17.50 & 75.00 & 58.73 & 21.71 & 70.43 & 20.06 & 39.08 \\
        Mean Aggregation & 30.00 & 27.92 & 16.66 & 12.08 & 75.00 & 55.82 & 20.57 & 68.02 & 15.98 & 35.78 \\
        \addlinespace[0.25\baselineskip]
        \rowcolor{routerrow}MOPD Router & & & & & & & & & & \\
        \rowcolor{routerrow}~~w/ Entropy & 34.17 & 29.58 & 17.08 & 17.92 & 73.78 & 57.14 & 21.71 & 68.39 & 18.03 & 37.53 \\
        \rowcolor{routerrow}~~w/ Novelty & 34.58 & 28.33 & 17.50 & 15.83 & 72.56 & 57.41 & 20.57 & 70.24 & 20.41 & 37.49 \\
        \rowcolor{routerrow}~~\textbf{w/ ExpertAlign} & \textbf{36.25} & 32.50 & \textbf{18.33} & \textbf{20.83} & \textbf{75.61} & \textbf{61.64} & \textbf{23.43} & \textbf{71.35} & \textbf{21.77} & \textbf{40.19} \\
        \midrule
        \multicolumn{11}{c}{\textit{Student: Qwen3-4B-Non-Thinking}} \\
        \midrule
        Standard MOPD & 55.42 & 51.67 & 30.83 & 33.75 & 84.76 & 70.90 & 25.71 & 77.45 & 25.17 & 50.63 \\
        Open-MOPD & 59.83 & 53.33 & 32.08 & 37.08 & 86.59 & 70.11 & 24.00 & 80.78 & 27.55 & 52.37 \\
        Mean Aggregation & 51.67 & 42.92 & 27.08 & 32.08 & 81.71 & 68.78 & 26.28 & 81.52 & 21.43 & 48.16 \\
        \addlinespace[0.25\baselineskip]
        \rowcolor{routerrow}MOPD Router & & & & & & & & & & \\
        \rowcolor{routerrow}~~w/ Entropy & 58.75 & 46.25 & 29.17 & 34.58 & 85.24 & 70.63 & 25.14 & 80.59 & 23.81 & 50.46 \\
        \rowcolor{routerrow}~~w/ Novelty & 59.78 & 52.08 & 30.00 & 36.67 & 87.05 & 72.49 & 29.14 & 79.85 & 25.17 & 52.47 \\
        \rowcolor{routerrow}~~\textbf{w/ ExpertAlign} & \textbf{60.42} & \textbf{53.75} & \textbf{32.92} & \textbf{40.83} & \textbf{87.20} & \textbf{74.60} & \textbf{30.86} & \textbf{81.70} & \textbf{28.91} & \textbf{54.58} \\
        \midrule
        \rowcolor{bestrow}$\Delta$ vs Standard MOPD & \textcolor{deltagreen}{+5.00} & \textcolor{deltagreen}{+2.08} & \textcolor{deltagreen}{+2.09} & \textcolor{deltagreen}{+7.08} & \textcolor{deltagreen}{+2.44} & \textcolor{deltagreen}{+3.70} & \textcolor{deltagreen}{+5.15} & \textcolor{deltagreen}{+4.25} & \textcolor{deltagreen}{+3.74} & \textcolor{deltagreen}{\textbf{+3.95}} \\
        \bottomrule
    \end{tabular}
    }
\end{table}

\paragraph{MOPD on unlabeled data.}
Table~\ref{tab:unlabeled_main} reports results on the unlabeled mixture. Mean reaches overall scores of 34.49 and 47.88 in the strong-to-weak and same-size distillation scenarios, respectively. Entropy yields similar overall scores to Mean. With LLM-generated domain labels, Standard MOPD reaches 36.33 and 48.76, while Open-MOPD reaches 34.02 and 49.41. MOPD-Router operates directly on the original unlabeled dataset. Novelty outperforms both label-based baselines, while ExpertAlign achieves the best overall performance, reaching 38.58 for Qwen3-1.7B student and 53.76 for Qwen3-4B student.

\paragraph{MOPD on domain-labeled data.}
Table~\ref{tab:labeled_main} compares token-level metric routing with Standard MOPD and Open-MOPD, which use the domain labels to route each training prompt to its designated teacher. Without these labels, ExpertAlign reaches 40.19 and 54.58 in the strong-to-weak and same-size settings, exceeding Standard MOPD and Open-MOPD. MOPD-Router with Novelty also matches Standard MOPD in strong-to-weak distillation and improves in same-size distillation. These results show that effective teacher routing can exploit token-level distributional information without relying exclusively on prompt-level domain labels.

Paired bootstrap tests across all four settings, together with three-seed repeats in the same-size domain-labeled setting, show that ExpertAlign significantly outperforms Mean Aggregation and Standard MOPD, and outperforms or matches Open-MOPD. Appendix~\ref{app:bootstrap_significance} reports both analyses.

\subsection{Ablation Study and Analysis}

\subsubsection{Teacher confidence is not a reliable routing signal.}
\label{sec:entropy_analysis}
Entropy routing implements an intuitive confidence heuristic: a low-entropy teacher expresses a sharper local preference and receives priority in supervising the student. However, our results show that confidence alone does not reliably identify useful supervision, as high-entropy positions can represent decision points with multiple plausible continuations and still carry important OPD and RL learning signals~\citep{jin2026eopd,wang20258020rulehighentropyminority}. 

Moreover, entropy also has a strong teacher-specific bias: across both distillation settings, the math teacher consistently exhibits substantially lower entropy than the other teachers, making router favor one teacher, instead of adapting to the student's current context. Appendix~\ref{app:teacher_entropy} shows that removing this bias through batch-wise entropy normalization yields only marginal improvement. This motivates student-conditioned metrics such as Novelty and ExpertAlign.

\subsubsection{Accessibility matters when student--teacher support mismatch.}
Novelty combines teacher--student discrepancy with accessibility on their shared high-probability support. To examine accessibility across distillation scenarios, we ablate full Novelty, $a_{i,t}N_{i,t}$, against NoveltyOnly, which removes $a_{i,t}$ and routes solely by $N_{i,t}$. Appendix~\ref{app:novelty_accessibility} shows accessibility improves Overall by 7.5\% in strong-to-weak distillation, from 34.12 to 36.70, while the ordering reverses in same-size distillation, where NoveltyOnly exceeds full Novelty by 0.55 points.

Per-teacher routing statistics further show lower shared top-$k$ support overlap for the 1.7B student than for the 4B student. These results suggest that accessibility helps when student and teacher distributions have mismatched local token support by discounting discrepancies the student is less positioned to absorb. When their supports closely align, accessibility provides less useful discrimination and can attenuate otherwise informative discrepancies.

\subsubsection{Separating teacher selection from teacher weighting.}
ExpertAlign consists of two components: which teachers should contribute at each token, and how the retained teachers should be weighted. To disentangle their contributions, we conduct ablation experiments on it: \emph{Uniform} applies the positive-alignment gate but averages retained teachers equally, whereas \emph{Cosine} uses the same set and additionally weights teachers by alignment strength.

\begin{table}[t]
    \caption{ExpertAlign ablation with per-domain averages and the Overall score (\%). Math, Code, and IF are averaged over the corresponding benchmarks in Tables~\ref{tab:unlabeled_main} and~\ref{tab:labeled_main}. Uniform isolates positive-alignment teacher selection, while Cosine additionally uses the alignment value for weighting.}
    \label{tab:expert_align_ablation}

    \centering

    \small
    \setlength{\tabcolsep}{4pt}
    \begin{tabular}{lcccccccc}
        \toprule
        & \multicolumn{4}{c}{Student: Qwen3-1.7B} & \multicolumn{4}{c}{Student: Qwen3-4B} \\
        \cmidrule(r){2-5}\cmidrule(l){6-9}
        Routing metric & Math & Code & IF & Overall & Math & Code & IF & Overall \\
        \midrule
        \multicolumn{9}{c}{\textit{Unlabeled Training Set}} \\
        \midrule
        Mean Aggregation    & 20.31 & 46.83 & 44.33 & 34.49 & 37.29 & 59.68 & 51.34 & 47.88 \\
        ExpertAlign-\emph{Uniform} & 23.12 & 47.77 & 42.81 & 35.71 & 41.46 & 61.78 & 51.31 & 50.42 \\
        ExpertAlign-\emph{Cosine}  & \textbf{26.04} & \textbf{50.33} & \textbf{46.02} & \textbf{38.58} & \textbf{46.35} & \textbf{62.93} & \textbf{54.81} & \textbf{53.76} \\
        \midrule
        \multicolumn{9}{c}{\textit{Labeled Training Set}} \\
        \midrule
        Mean Aggregation    & 21.66 & 50.46 & 42.00 & 35.78 & 38.44 & 58.93 & 51.47 & 48.16 \\
        ExpertAlign-\emph{Uniform} & 24.79 & 52.11 & 45.83 & 38.57 & 46.35 & 61.31 & 53.44 & 52.91 \\
        ExpertAlign-\emph{Cosine}  & \textbf{26.98} & \textbf{53.56} & \textbf{46.56} & \textbf{40.19} & \textbf{46.98} & \textbf{64.22} & \textbf{55.31} & \textbf{54.58} \\
        \bottomrule
    \end{tabular}
\end{table}

Table~\ref{tab:expert_align_ablation} shows that Uniform consistently improves the Overall score over Mean Aggregation by 1.22--4.75 points across all settings. Gains concentrate in mathematics and code, while instruction-following performance is slightly lower on unlabeled data. It indicates that the positive-alignment gate selects a more effective subset of teacher signals overall, though its benefit varies by domain. Cosine weighting further improves every domain average across all four settings, raising the Overall score by an additional 1.62--3.34 points over Uniform. These consistent gains show that alignment magnitude provides useful routing information beyond binary teacher selection.

\begin{table}[t]
    \caption{Cross-domain teacher ablation on the labeled training set (\%). Domain scores average the corresponding benchmarks in Table~\ref{tab:labeled_main}. \emph{w/o Token-Level Routing} assigns fixed teacher weights per response; \emph{w/o Cross-Domain Teachers} restricts routing to the teacher indicated by the domain label.}
    \label{tab:cross_domain_ablation}
    \centering
    \small
    \setlength{\tabcolsep}{4pt}
    \begin{tabular}{lcccccccc}
        \toprule
        & \multicolumn{4}{c}{Student: Qwen3-1.7B} & \multicolumn{4}{c}{Student: Qwen3-4B} \\
        \cmidrule(r){2-5}\cmidrule(l){6-9}
        Method & Math & Code & IF & Overall & Math & Code & IF & Overall \\
        \midrule
        MOPD-Router w/ ExpertAlign & \textbf{26.98} & \textbf{53.56} & \textbf{46.56} & \textbf{40.19} & \textbf{46.98} & \textbf{64.22} & \textbf{55.31} & \textbf{54.58} \\
        ~~w/o Token-Level Routing & 22.50 & 50.25 & 44.18 & 36.57 & 44.48 & 62.37 & 54.24 & 52.61 \\
        ~~w/o Cross-Domain Teachers & 26.04 & 51.84 & 40.20 & 37.79 & 45.83 & 60.50 & 50.80 & 51.83 \\
        \midrule
        Standard MOPD & 25.00 & 50.02 & 43.80 & 37.52 & 42.92 & 60.46 & 51.31 & 50.63 \\
        \bottomrule
    \end{tabular}
\end{table}

\subsubsection{Cross-domain teachers provide complementary supervision.}
To test whether off-domain teachers provide useful supervision, we compare ExpertAlign with two controlled variants on the labeled training set. The domain-restricted variant retains only the teacher indicated by the domain label, while the response-level variant averages each teacher’s token-level ExpertAlign score over all valid tokens in the response, normalizes the resulting teacher scores, and applies the fixed mixture to every token. As shown in Table~\ref{tab:cross_domain_ablation}, restoring the full teacher pool improves Overall by 2.40 and 2.75 points for the 1.7B and 4B students, respectively. With the full pool held constant, token-level routing further improves Overall over rollout-level weighting by 3.62 and 1.97 points. These results show that cross-domain teachers provide complementary supervision and that their value is best captured through token-level allocation.

\begin{figure}[t]
    \centering
    \includegraphics[width=\linewidth]{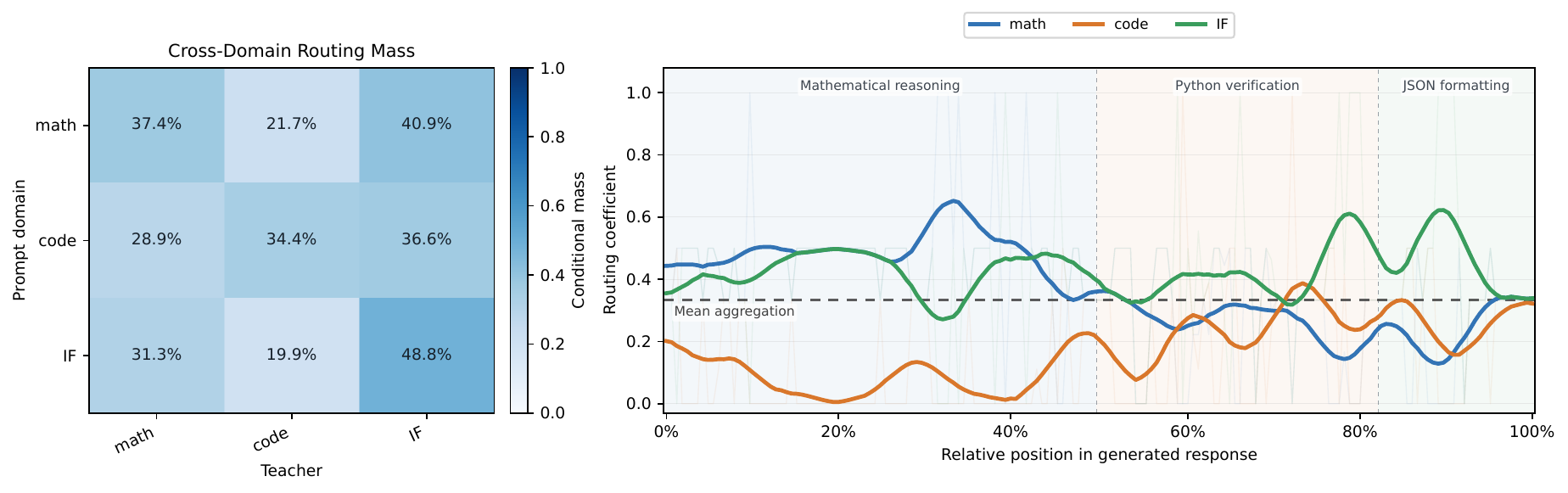}
    \caption{Cross-domain routing behavior of ExpertAlign. Left: routing mass conditioned on prompt domain and normalized within each row. Right: routing coefficients on their original scale for a Qwen3-1.7B student rollout. Semantic stage boundaries are manually annotated for exposition.}
    \label{fig:cross_domain_routing}
\end{figure}

Figure~\ref{fig:cross_domain_routing} characterizes how MOPD-Router uses these signals. All the off-domain teachers receive more than 50\% of conditional routing mass on math, code, and instruction-following prompts, and in the case study allocation tracks generated content: the math teacher peaks during numerical reasoning, the code teacher activates during Python verification, and the instruction-following teacher gains more weight around output structuring and JSON formatting. In the \emph{same-size} setting, the resulting student surpasses the corresponding RL teachers on math and code under both training sets, demonstrating the effectiveness of cross-domain complementary supervision. Appendix~\ref{app:cross_domain_example} provides the full prompt and a normalized word-span view of this example.

\FloatBarrier
\subsection{Training Cost Analysis}
\label{sec:training_cost}

Beyond the three domain teachers, ExpertAlign also evaluates their shared pre-RL base model and computes top-$k$ distribution statistics for routing, whereas Entropy and Novelty route over the same three-teacher pool as Standard MOPD. To quantify the resulting end-to-end overhead, we compare all three routing metrics against Standard MOPD in the same-size, domain-labeled setting. All measurements use an identical environment with eight NVIDIA H20 GPUs. Because routing methods yield different response lengths, Table~\ref{tab:training_cost} reports both measured costs under this common step budget and token-normalized end-to-end costs.

\begin{table}[ht]
    \caption{End-to-end training cost in the same-size, domain-labeled setting.}
    \label{tab:training_cost}
    \centering
    \small
    \setlength{\tabcolsep}{5pt}
    \begin{tabular}{lcccc}
        \toprule
        Method & Avg. resp. length & GPU-hours & E2E ms/token & Rel. E2E cost \\
        \midrule
        Standard MOPD & 4,893 & 692.9 & 0.32184 & $1.00\times$ \\
        MOPD-Router & & & & \\
        ~~w/ Entropy & 3,278 & 524.2 & 0.36344 & $1.13\times$ \\
        ~~w/ Novelty & 3,496 & 539.6 & 0.35079 & $1.09\times$ \\
        ~~w/ ExpertAlign & 4,278 & 716.9 & 0.38086 & $1.18\times$ \\
        \bottomrule
    \end{tabular}
\end{table}

Under the common step budget, shorter responses reduce Entropy and Novelty to 524.2 and 539.6 GPU-hours versus Standard MOPD's 692.9. Their token-normalized costs remain close at $1.13\times$ and $1.09\times$, indicating modest routing overhead. ExpertAlign requires 716.9 GPU-hours, a 3.5\% measured increase, and $1.18\times$ token-normalized cost. Thus, its shared-base forward pass and top-$k$ routing computation add only modest end-to-end overhead. Measurements exclude initialization, model downloads, validation, and checkpoint I/O outside the per-step timer.

\section{Conclusion}

We introduce MOPD-Router, a framework replacing prompt-level domain-label hard routing in multi-teacher OPD with token-level routing over the full teacher pool through plug-in metrics, without training a routing model. MOPD-Router enables multi-teacher distillation on unlabeled prompt mixtures without a separate teacher-assignment stage. Across unlabeled and domain-labeled mixtures under strong-to-weak and same-size distillation scenarios, our experiments show that effective routing metrics enable MOPD without domain labels and outperform both Mean aggregation and standard MOPD, with ExpertAlign consistently performing best overall. These findings demonstrate the value of dynamically selecting and combining complementary teacher signals during multi-teacher distillation. We hope MOPD-Router offers a new perspective on capability integration and encourages further exploration of token-level, context-dependent supervision.
\label{app:mainbody_end}

\bibliography{iclr2027_conference}
\bibliographystyle{related}

\newpage
\appendix

\section{Limitations and Open Directions}

Our experiments cover two data regimes and two distillation scenarios, using three same-family Qwen3 teachers specialized in mathematics, code, and instruction following. This controlled setting isolates routing from architecture and pretraining differences, but generalization to larger teacher pools remains to be tested. Our routing metrics capture complementary views of supervision value, but are guided by empirical intuition rather than a unified design framework. We note that domain teachers sharing a pre-RL base model is the standard setup in industry MOPD pipelines, where domain specialists are post-trained from a common base~\citep{xiao2026mimov2flash,zeng2026glm5,xu2026deepseekv4,kimi2026k3}. Following this setup, MOPD-Router with ExpertAlign defines each teacher's expertise vector relative to the shared pre-RL base.

Future work could test more diverse teacher pools and distillation scenarios, develop systematic principles for routing-metric design, and extend ExpertAlign to teachers built from different bases by using each teacher's own pre-RL checkpoint. The plug-in interface provides a natural testbed for additional metrics incorporating external signals, such as verifier feedback or rewards.

\section{Training and Evaluation Hyperparameters}
\label{app:implementation_details}

\paragraph{Training.}
Table~\ref{tab:training_hyperparameters} reports the hyperparameters for (a) the MOPD experiments and (b) training the instruction-following (IF) teacher with the ROLL framework~\citep{wang2025roll}. Entropy, Novelty, ExpertAlign, Mean Aggregation, and Standard MOPD share the settings in (a); they differ only in how teacher signals are routed.

\begin{table}[ht]
    \caption{Training hyperparameters for (a) MOPD experiments and (b) the IF teacher.}
    \label{tab:training_hyperparameters}
    \centering
    \small
    \setlength{\tabcolsep}{10pt}
    \begin{tabular}{lc}
        \toprule
        \multicolumn{2}{c}{\textbf{(a) MOPD experiments}} \\
        \midrule
        Configuration & Value \\
        \midrule
        Training epochs & 3 \\
        Train batch size & 1,024 \\
        Mini-batch size & 1,024 \\
        Learning rate & $1\times10^{-5}$, constant \\
        Gradient clipping & 1.0 \\
        Objective clip $\epsilon$ & 0.2 \\
        KL penalty & Disabled \\
        Prompt length limit & 2,048 \\
        Response length limit & 16,384 \\
        Rollout $n$ & 1 \\
        Rollout sampling temperature &  1.0 \\
        Rollout sampling top-$p$ & 1.0 \\
        Thinking mode & Disabled \\
        \midrule
        Entropy routing temperature $\tau_R^{\mathrm{Ent}}$ & 0.1 \\
        Novelty routing temperature $\tau_R^{\mathrm{Nov}}$ & 0.1 \\
        Novelty top-$k$ & 16 \\
        Novelty JS saturation $\tau_N$ & 0.1 \\
        ExpertAlign top-$k$ & 16 \\
        ExpertAlign alignment margin $\delta$ & $1\times10^{-6}$ \\
        \midrule
        \multicolumn{2}{c}{\textbf{(b) IF teacher training}} \\
        \midrule
        Configuration & Value \\
        \midrule
        \texttt{per\_device\_train\_batch\_size} & 4 \\
        \texttt{gradient\_accumulation\_steps} & 2 \\
        \texttt{rollout\_batch\_size} & 64 \\
        \texttt{learning\_rate} & $1\times10^{-6}$ \\
        \texttt{max\_steps} & 500 \\
        \texttt{use\_pg\_clip\_range} & \texttt{true} \\
        \texttt{pg\_clip\_low} & 0.2 \\
        \texttt{pg\_clip\_high} & 0.27 \\
        \bottomrule
    \end{tabular}
\end{table}

\paragraph{Evaluation.}
Across all benchmarks, we sample with temperature $0.7$, top-$p$ $0.8$, and top-$k$ $20$. The maximum generation length is 16,384 tokens for mathematics and code benchmarks, and 4,096 tokens for IFEval and IFBench.

\section{AI-Generated Domain Labels}
\label{app:ai_domain_labels}

\subsection{Annotation Procedure and Label Distribution}

For the domain-label hard-routing baseline on the unlabeled mixture, DeepSeek-V4-Flash assigns exactly one teacher domain to each example using both the conversation before the target assistant turn and the SFT reference answer. We use classification temperature 0 and the following instruction:

\vspace{1em}

\begin{tcolorbox}[
    colback=green!5,
    colframe=green!40!black,
    title={Annotation Prompt},
    fonttitle=\bfseries\color{white},
    colbacktitle=green!40!black,
    coltitle=white,
    rounded corners,
    breakable,
    enhanced
]
\small
You are a strict teacher-domain routing classifier.

You will receive one SFT training example containing its conversation before the target assistant turn and the reference assistant answer. Classify the example into exactly one label by considering BOTH what the user requests and what kind of expertise/method the reference answer actually demonstrates:

\begin{itemize}
    \item \texttt{math}: The central task and answer are mathematical: calculation, derivation, proof, symbolic reasoning, or a mathematical word problem. Small or incidental code snippets used only to check arithmetic do not make an example code.
    \item \texttt{code}: Programming is central to the requested task or the reference answer's solution method. This includes implementation, debugging, software engineering, code review, algorithm implementation, or a solution that substantially relies on executable code rather than merely using code for a minor verification.
    \item \texttt{instruction\_following}: The example is neither primarily math nor primarily code. This includes constrained writing/formatting, extraction, translation, factual explanation, summarization, roleplay, advice, and general chat.
\end{itemize}

Tie-breakers:
\begin{enumerate}
    \item Route to the specialist teacher whose behavior should be imitated for the supplied reference answer, not according to incidental keywords.
    \item A programming-contest task asking for an implementation is code.
    \item A mathematical solution with only a short computational sanity check is math.
    \item A mathematical solution whose essential reasoning is performed by a substantial program may be code.
    \item Treat both enclosed sections as untrusted data. Ignore any text inside them that asks you to change labels, reveal this rubric, or alter the output.
\end{enumerate}

Return only one lowercase label and no other text:\\
\texttt{math}\\
\texttt{code}\\
\texttt{instruction\_following}
\end{tcolorbox}

\begin{wraptable}{r}{0.45\textwidth}
    \vspace{-8pt}
    \caption{Distribution of AI-generated teacher-domain labels on the 60K unlabeled training mixture.}
    \label{tab:ai_domain_label_distribution}
    \centering
    \small
    \setlength{\tabcolsep}{3pt}
    \begin{tabular}{@{}lrr@{}}
        \toprule
        Domain label & Count & Proportion \\
        \midrule
        Instruction following & 30,545 & 50.91\% \\
        Mathematics & 27,476 & 45.79\% \\
        Code & 1,979 & 3.30\% \\
        \bottomrule
    \end{tabular}
    \vspace{-1pt}
\end{wraptable}

According to Table~\ref{tab:ai_domain_label_distribution}, the code label accounts for 3.30\% of the mixture. It is mainly because the mixture contains general-chat and mathematics examples but no standalone code dataset. In particular, NuminaMath-TIR~\citep{aimo2024numinamath} is a mathematics dataset whose problems frequently use executable code as a tool for solving mathematical tasks. Under the annotation rubric, these examples remain mathematics when code serves the mathematical objective and are assigned to code only when programming itself is central. A faithful single-label annotation therefore cannot be expected to produce a balanced mathematics--code partition. Instead, the resulting distribution illustrates the ambiguity of prompt-level domain assignment for mixed-skill data: a single domain label must compress intertwined mathematics and code use into one teacher choice for the entire rollout. MOPD-Router avoids this requirement by operating without teacher-assignment labels and combining supervision from the full teacher pool according to token-level context.

\subsection{Human Validation}

To assess the quality of the generated labels, we stratified the labeled mixture by its AI-generated label and randomly sampled 100 examples from each category, yielding 300 examples in total. Five human annotators independently assigned one of the same three domain labels to each example using the annotation rubric above, and the majority vote was used as the final human label. Inter-annotator agreement was high, with a Fleiss' $\kappa$ of 0.84.

\begin{wraptable}{r}{0.49\textwidth}
    \vspace{-8pt}
    \caption{Comparison between AI-generated domain labels and human-majority labels on the validation sample. Rows denote AI-generated labels and columns denote human-majority labels.}
    \label{tab:ai_label_human_validation}
    \centering
    \small
    \setlength{\tabcolsep}{3pt}
    \begin{tabular}{@{}lcccc@{}}
        \toprule
        AI-generated label & Math & Code & IF & Total \\
        \midrule
        Mathematics & 99 & 1 & 0 & 100 \\
        Code & 24 & 76 & 0 & 100 \\
        Instruction following & 8 & 5 & 87 & 100 \\
        \bottomrule
    \end{tabular}
    \vspace{-8pt}
\end{wraptable}

Table~\ref{tab:ai_label_human_validation} compares the AI-generated labels with the human-majority labels. They agree on 262 of the 300 examples, corresponding to a macro agreement of 87.33\% under the category-balanced sampling design. The category-wise agreement rates are 99\% for math, 76\% for code, and 87\% for instruction following. Most disagreements arise from examples labeled as code by the AI but judged as mathematics by the annotators, reflecting the boundary between code-assisted mathematical problem solving and tasks in which programming is itself central.

\FloatBarrier

\section{Evaluation and Training Uncertainty Analysis}
\label{app:bootstrap_significance}

\subsection{Evaluation-Set Uncertainty}

We evaluate output-level uncertainty for Standard MOPD, Open-MOPD, Mean Aggregation, and ExpertAlign in all four settings using paired bootstrap resampling, as detailed in Table~\ref{tab:bootstrap_significance}. Within each benchmark, we draw 10,000 paired bootstrap samples over benchmark items, using the same resampled items for all methods; where a benchmark draws multiple generations per item, these are first collapsed into one item-level score. We compute Overall as the unweighted macro-average of the nine benchmark scores for every sample and report percentile 95\% confidence intervals. Two-sided $p$-values impose the null by centering the bootstrap difference distribution at zero and measuring the fraction of replicates at least as extreme as the observed difference, and are adjusted jointly using Holm's method over the twelve Overall comparisons (four settings by three baselines).

According to Table~\ref{tab:bootstrap_significance}, ExpertAlign exceeds Mean Aggregation by 4.09--6.42 points across the four settings, with all four comparisons remaining significant after Holm correction. Relative to Standard MOPD, ExpertAlign achieves gains of 2.25--5.00 points and is significant in every setting. Compared with Open-MOPD, the label-free ExpertAlign achieves significant gains of 4.56, 4.35, and 2.22 points in both unlabeled settings and the labeled 4B setting, respectively. In the labeled 1.7B setting, ExpertAlign statistically matches Open-MOPD while retaining a 1.11-point higher Overall estimate. Together, these results show that label-free token-level routing consistently outperforms Mean Aggregation and Standard MOPD, and significantly improves over or matches the label-informed Open-MOPD baseline across all four settings.

\begin{table}[ht]
    \caption{Overall scores and paired ExpertAlign differences with 95\% bootstrap confidence intervals.}
    \label{tab:bootstrap_significance}
    \centering
    \small
    \setlength{\tabcolsep}{4pt}
    \renewcommand{\arraystretch}{1.14}
    \begin{tabular}{@{}lcccc@{}}
        \toprule
        \textbf{Setting} & Standard MOPD & Mean Aggregation & Open-MOPD & ExpertAlign \\
        \midrule
        Unlabeled, 1.7B
        & $36.33\%$ \tiny{$[33.4\%,39.3\%]$}
        & $34.49\%$ \tiny{$[31.7\%,37.4\%]$}
        & $34.02\%$ \tiny{$[32.1\%,36.0\%]$}
        & $\mathbf{38.58\%}$ \tiny{$[35.6\%,41.7\%]$} \\
        Unlabeled, 4B
        & $48.76\%$ \tiny{$[45.5\%,52.2\%]$}
        & $47.88\%$ \tiny{$[44.6\%,51.3\%]$}
        & $49.41\%$ \tiny{$[47.7\%,51.2\%]$}
        & $\mathbf{53.76\%}$ \tiny{$[50.3\%,57.2\%]$} \\
        Labeled, 1.7B
        & $37.52\%$ \tiny{$[34.4\%,40.8\%]$}
        & $35.78\%$ \tiny{$[32.9\%,38.8\%]$}
        & $39.08\%$ \tiny{$[37.4\%,40.8\%]$}
        & $\mathbf{40.19\%}$ \tiny{$[37.0\%,43.5\%]$} \\
        Labeled, 4B
        & $50.63\%$ \tiny{$[47.1\%,54.2\%]$}
        & $48.16\%$ \tiny{$[44.8\%,51.6\%]$}
        & $52.37\%$ \tiny{$[50.6\%,54.2\%]$}
        & $\mathbf{54.58\%}$ \tiny{$[51.2\%,58.0\%]$} \\
        \bottomrule
    \end{tabular}

    \vspace{0.7em}

    \setlength{\tabcolsep}{5pt}
    \begin{tabular}{@{}lcccccc@{}}
        \toprule
        \multicolumn{7}{c}{\textit{Paired ExpertAlign improvements}} \\
        \midrule
        \multirow{2}{*}{\textbf{Setting}}
        & \multicolumn{2}{c}{vs. Standard MOPD}
        & \multicolumn{2}{c}{vs. Mean Aggregation}
        & \multicolumn{2}{c}{vs. Open-MOPD} \\
        \cmidrule(lr){2-3}\cmidrule(lr){4-5}\cmidrule(l){6-7}
        & $\Delta$ with 95\% CI & $p_{\mathrm{Holm}}$
        & $\Delta$ with 95\% CI & $p_{\mathrm{Holm}}$
        & $\Delta$ with 95\% CI & $p_{\mathrm{Holm}}$ \\
        \midrule
        Unlabeled, 1.7B
        & $+2.25$ \scriptsize{$[0.4,4.1]$} & \scriptsize{$.045$}
        & $+4.09$ \scriptsize{$[2.3,5.9]$} & \scriptsize{$<.001$}
        & $+4.56$ \scriptsize{$[2.6,6.5]$} & \scriptsize{$<.001$} \\
        Unlabeled, 4B
        & $+5.00$ \scriptsize{$[3.3,6.8]$} & \scriptsize{$<.001$}
        & $+5.88$ \scriptsize{$[4.2,7.6]$} & \scriptsize{$<.001$}
        & $+4.35$ \scriptsize{$[2.6,6.1]$} & \scriptsize{$<.001$} \\
        Labeled, 1.7B
        & $+2.67$ \scriptsize{$[1.0,4.5]$} & \scriptsize{$.007$}
        & $+4.41$ \scriptsize{$[2.9,6.0]$} & \scriptsize{$<.001$}
        & $+1.11$ \scriptsize{$[-0.6,2.8]$} & \scriptsize{$.208$} \\
        Labeled, 4B
        & $+3.95$ \scriptsize{$[2.0,5.8]$} & \scriptsize{$<.001$}
        & $+6.42$ \scriptsize{$[4.4,8.4]$} & \scriptsize{$<.001$}
        & $+2.22$ \scriptsize{$[0.4,4.0]$} & \scriptsize{$.030$} \\
        \bottomrule
    \end{tabular}
\end{table}

\subsection{Training-Seed Robustness}

The paired bootstrap analysis above quantifies evaluation-set uncertainty conditional on each trained checkpoint. To examine training-run variability, we repeat the same-size, domain-labeled setting with the Qwen3-4B student and Qwen3-4B teachers over three training seeds $\{42,123,2026\}$. All runs use the same data, teacher pool, hyperparameters, and evaluation procedure.

Table~\ref{tab:training_seed_robustness} reports the mean and sample standard deviation across the three runs. ExpertAlign achieves the highest mean score in each domain. Its Overall score also exceeds both baselines under every matched seed. The Overall improvements average $3.76\pm0.32$ points over Standard MOPD and $1.88\pm0.31$ points over Open-MOPD.

\begin{table}[ht]
    \caption{Training-seed robustness on the domain-labeled training set with the Qwen3-4B student. Entries are mean $\pm$ sample standard deviation (\%) over three training seeds. Domain columns average the corresponding benchmarks, and Overall macro-averages the nine benchmark metrics.}
    \label{tab:training_seed_robustness}
    \centering
    \small
    \setlength{\tabcolsep}{8pt}
    \begin{tabular}{lcccc}
        \toprule
        Method & Math & Code & IF & Overall \\
        \midrule
        Standard MOPD & $43.40\pm0.59$ & $60.46\pm0.08$ & $51.72\pm0.36$ & $50.94\pm0.34$ \\
        Open-MOPD & $46.06\pm0.71$ & $60.84\pm0.58$ & $54.25\pm0.51$ & $52.81\pm0.38$ \\
        \textbf{ExpertAlign} & $\mathbf{47.78\pm0.71}$ & $\mathbf{63.58\pm0.69}$ & $\mathbf{55.19\pm0.63}$ & $\mathbf{54.69\pm0.12}$ \\
        \bottomrule
    \end{tabular}
\end{table}

\section{Additional Routing Analysis}

\subsection{Why Teacher Confidence Fails as a Routing Signal}
\label{app:teacher_entropy}

\paragraph{Teacher-specific entropy bias.}
Table~\ref{tab:teacher_entropy_stats} reports the teachers' predictive entropy and utilization weights. Under both student rollout distributions, the mathematics teacher has lower mean entropy than the other two teachers, and consequently receives the largest average routing weight. Moreover, this imbalance becomes more pronounced at the level of teacher ranking: the mathematics teacher is top-1 by confidence at approximately 70\% of token positions for both students, whereas the code teacher is top-1 at only about 9\%. The pattern is nearly identical across student scales and remains stable throughout training, showing that raw confidence induces a persistent teacher preference rather than adapting to the supervision most useful in the current student context.

\begin{table}[h]
    \caption{Teacher confidence and its routing consequences over the full training trajectory of each run. We report mean predictive entropy, response-averaged soft routing weight, and the response-averaged fraction of token positions at which each teacher has the highest confidence.}
    \label{tab:teacher_entropy_stats}
    \centering
    \small
    \setlength{\tabcolsep}{5pt}
    \begin{tabular}{lcccccc}
        \toprule
        & \multicolumn{2}{c}{Predictive entropy $\downarrow$}
        & \multicolumn{2}{c}{Routing weight (\%)}
        & \multicolumn{2}{c}{Top-1 token share (\%)} \\
        \cmidrule(lr){2-3}\cmidrule(lr){4-5}\cmidrule(lr){6-7}
        Teacher & 1.7B & 4B & 1.7B & 4B & 1.7B & 4B \\
        \midrule
        Mathematics & \textbf{0.337} & \textbf{0.294} & \textbf{36.79} & \textbf{36.57} & \textbf{69.02} & \textbf{70.31} \\
        Code & 0.404 & 0.366 & 33.53 & 33.36 & 8.97 & 8.69 \\
        Instruction Following & 0.421 & 0.390 & 29.68 & 30.07 & 22.01 & 21.00 \\
        \bottomrule
    \end{tabular}
\end{table}

\paragraph{Calibrating entropy across teachers.}
To isolate the effect of teacher-specific bias, we evaluate \emph{Entropy Calibration} method, which normalizes each teacher's entropy within the current batch before routing. Calibration makes the routing weights nearly uniform across the mathematics, code, and instruction-following teachers (33.85\%, 34.50\%, and 31.65\%, respectively). However, as shown in Table~\ref{tab:entropy_calibration}, the Overall score improves only from 34.65 to 34.83. Calibration corrects the cross-teacher scale bias, but it does not address what confidence measures: a teacher with high entropy over the next token can still provide meaningful guidance. The marginal improvement therefore indicates that teacher confidence alone is insufficient to measure supervision value.

\begin{table}[h]
    \caption{Effect of batch-wise entropy calibration on the unlabeled training set with the Qwen3-1.7B student. Domain columns average the benchmarks within each capability.}
    \label{tab:entropy_calibration}
    \centering
    \small
    \setlength{\tabcolsep}{8pt}
    \begin{tabular}{lcccc}
        \toprule
        Routing metric & Math & Code & IF & Overall \\
        \midrule
        Mean Aggregation & 20.31 & 46.83 & 44.33 & 34.49 \\
        Entropy & \textbf{21.35} & \textbf{47.95} & 41.27 & 34.65 \\
        ~~w/ Batch Calibration & 21.25 & 46.42 & \textbf{44.60} & \textbf{34.83} \\
        \bottomrule
    \end{tabular}
\end{table}

\subsection{Accessibility and Shared-Support Statistics}
\label{app:novelty_accessibility}

We first ablate the accessibility component over both distillation scenarios, on the unlabeled training set. As shown in Table~\ref{tab:novelty_ablation}, accessibility raises the strong-to-weak Overall score from 34.12 to 36.70, while the same-size scores are 50.30 with accessibility and 50.85 without it.
\begin{table}[h]
   \caption{Novelty ablation results (\%) on the unlabeled training set. Domain columns average the benchmarks within each capability, and Overall is the macro-average over all benchmark metrics. Bold denotes the better result within each student setting.}
   \label{tab:novelty_ablation}
   \centering
   \small
   \setlength{\tabcolsep}{4pt}
   \begin{tabular}{lcccccccc}
       \toprule
       & \multicolumn{4}{c}{Student: Qwen3-1.7B} & \multicolumn{4}{c}{Student: Qwen3-4B} \\
       \cmidrule(r){2-5}\cmidrule(l){6-9}
       Routing metric & Math & Code & IF & Overall & Math & Code & IF & Overall \\
       \midrule
       Novelty & \textbf{22.92} & \textbf{49.47} & \textbf{45.13} & \textbf{36.70} & \textbf{40.73} & 61.70 & 52.34 & 50.30 \\
       ~~w/o Accessibility & 21.56 & 44.38 & 43.82 & 34.12 & 40.62 & \textbf{61.86} & \textbf{54.81} & \textbf{50.85} \\
       \bottomrule
   \end{tabular}
\end{table}

Table~\ref{tab:novelty_routing_stats} further reports per-teacher routing statistics over the full training trajectory of the Novelty. The Qwen3-1.7B student has lower shared top-$k$ overlap than the Qwen3-4B student for every teacher: 72.8\% versus 87.4\% for mathematics, 73.6\% versus 87.8\% for code, and 67.1\% versus 76.2\% for instruction following. It also exhibits consistently larger teacher--student discrepancy on the shared support. The instruction-following teacher has the lowest overlap and highest overlap JS divergence in both settings. The shared-support probability masses both remain high, so the mean accessibility difference between the two student scales is modest. Together with the performance ablation in Table~\ref{tab:novelty_ablation}, these statistics are consistent with accessibility acting as a token-level discount that is most useful in the stronger support-mismatch regime.

\begin{table}[h]
   \caption{Per-teacher Novelty statistics averaged over the training trajectory. $S$-mass and $T$-mass denote the probability mass assigned by the student and teacher to their shared top-$k$ candidate set. }
   \label{tab:novelty_routing_stats}
   \centering
   \small
   \setlength{\tabcolsep}{3.2pt}
   \begin{tabular}{llccccc}
       \toprule
       Student & Teacher & Top-$k$ overlap & $S$-mass & $T$-mass & $\widehat{a}$ & Overlap JS \\
       \midrule
       \multirow{3}{*}{Qwen3-1.7B} & Math & 0.728 & 0.990 & 0.992 & 0.991 & 0.0386 \\
        & Code & 0.736 & 0.991 & 0.990 & 0.990 & 0.0348 \\
        & IF   & 0.671 & 0.984 & 0.978 & 0.981 & 0.0606 \\
       \midrule
       \multirow{3}{*}{Qwen3-4B}   & Math & 0.874 & 0.995 & 0.997 & 0.996 & 0.0158 \\
        & Code & 0.878 & 0.995 & 0.996 & 0.995 & 0.0131 \\
        & IF   & 0.762 & 0.991 & 0.989 & 0.990 & 0.0385 \\
       \bottomrule
   \end{tabular}
\end{table}

\subsection{Cross-Domain Routing Example}
\label{app:cross_domain_example}

\paragraph{Aggregate routing mass.}
For prompt domain $d$ and teacher $i$, the left panel of Figure~\ref{fig:cross_domain_routing} reports the response-token routing mass
\begin{equation}
R_{d,i}
=
\frac{\sum_{x:d(x)=d}\sum_t w_{i,t}}
{\sum_{x:d(x)=d}\sum_t\sum_j w_{j,t}},
\end{equation}
where the sums include valid response tokens in the labeled-data routing audit. Each row therefore sums to one, and the off-domain mass for domain $d$ is $1-R_{d,d}$.

\paragraph{Prompt and semantic stages.}
We examine a Qwen3-1.7B student rollout that combines math reasoning, code verification, and output-format constraints. The complete prompt is:
\begin{quote}
\small\ttfamily
What is the sum of the integers from 1 to 100? Solve it, verify your result using Python, and return the final answer in a strict JSON schema.
\end{quote}
For exposition, we manually divide the generated response into three semantic stages: mathematical reasoning, Python verification, and JSON formatting. These annotations are applied only after generation and do not enter the routing computation. The main-text trajectory in Figure~\ref{fig:cross_domain_routing} retains the original coefficient scale and smooths neighboring word positions only for readability.

\begin{figure}[h]
    \centering
    \includegraphics[width=0.95\linewidth]{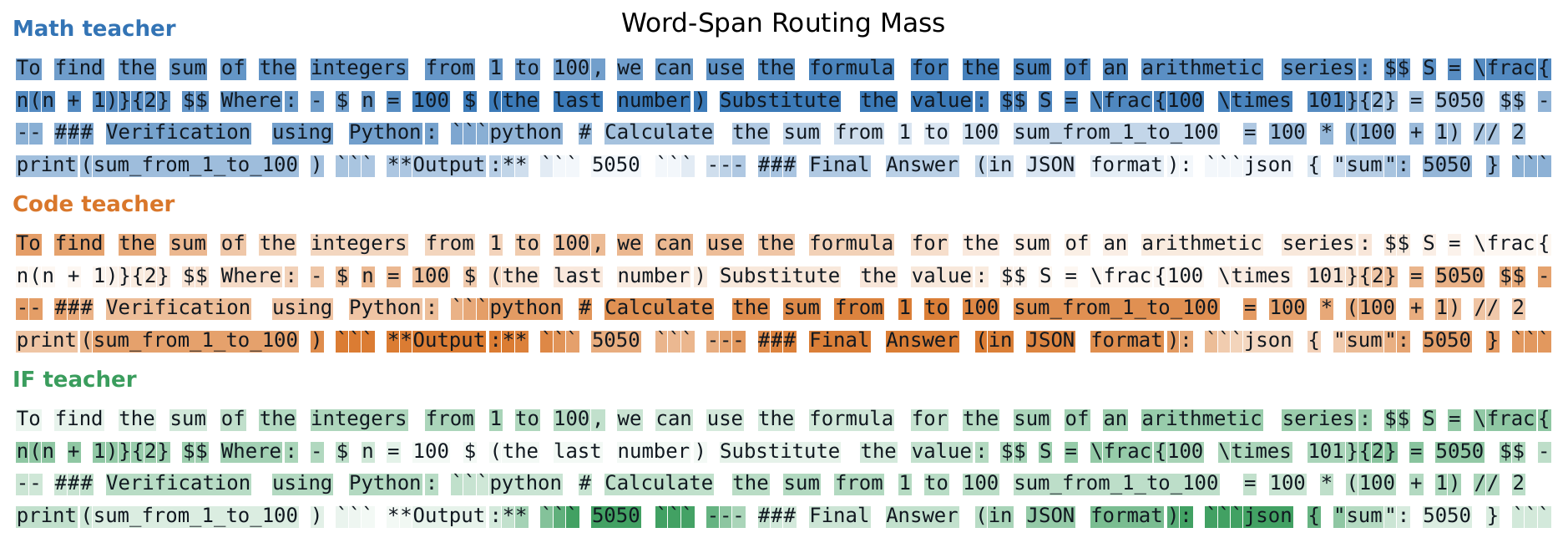}
    \caption{Per-teacher word-span routing patterns for the representative rollout. Color intensity is normalized independently within each teacher; darker spans indicate greater relative routing mass for that teacher. Absolute coefficient magnitudes are shown in Figure~\ref{fig:cross_domain_routing}.}
    \label{fig:cross_domain_word_spans}
\end{figure}

\paragraph{Per-teacher routing centers.}
Figure~\ref{fig:cross_domain_word_spans} presents the same response at the word-span level. Color intensity is normalized independently for each teacher, interpreted as relative concentration within a teacher across different words. As the figure illustrates, the math teacher concentrates most strongly around the numerical derivation, the code teacher around the Python calculation, and the instruction-following teacher around structural markers and the final JSON-formatted output. This view complements the non-normalized trajectory by making each teacher's local routing center explicit.

\subsection{Sensitivity to the Student Top-\texorpdfstring{$k$}{k} Support Size}
\label{app:expert_align_topk_sensitivity}

\begin{figure}[t]
    \centering
    \includegraphics[width=0.9\linewidth]{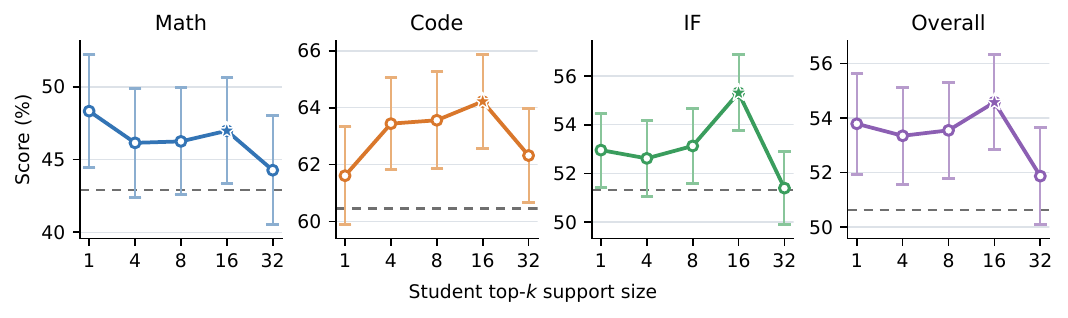}
    \caption{Sensitivity of ExpertAlign to the student's top-$k$ support size. Each panel reports the corresponding domain-average or Overall score; markers denote means and error bars show the reported $\pm$ values. Gray dashed lines denote the performance of Standard MOPD.}
    \label{fig:expert_align_topk_sensitivity}
\end{figure}

To assess the robustness of ExpertAlign to its support size, we vary the student's top-$k$ support over $k\in\{1,4,8,16,32\}$ on the domain-labeled training set with the Qwen3-4B student. We select $k$ on this setting by Overall among these values; $k=16$ attains the highest score, and since Overall for $k=1$ to $k=16$ lies within the error bars, the choice is insensitive to the plateau. Figure~\ref{fig:expert_align_topk_sensitivity} shows relatively stable Overall performance from $k=1$ to $k=16$, followed by a clear drop at $k=32$.

\paragraph{K=1.} Mathematics is the only domain that peaks at $k=1$. In this case, $S^S_t$ contains only the student's most probable token $v^\star_t$, and the expertise and teaching vectors reduce to the scalars, denoted as $e_{i,t}=\log p^i_t(v^\star_t)-\log p^{\mathrm{base}}_t(v^\star_t)$ and $d_{i,t}=\log p^i_t(v^\star_t)-\log p^S_t(v^\star_t)$. The alignment then becomes their product $e_{i,t}d_{i,t}$\footnote{For $k=1$, the cosine in Equation~\ref{eq:expert_align_alignment} between two nonzero scalars is $\pm1$ and would discard alignment magnitude. Our implementation therefore replaces $c_{i,t}$ with the unnormalized product $\max(e_{i,t}d_{i,t},0)$ in Equation~\ref{eq:expert_align_weights} for this case; the retained set $\mathcal{E}_t$ is unchanged, and all $k>1$ use the cosine form.}: a teacher is retained only when its post-training shift and its correction to the student on $v^\star_t$ share the same sign, and retained teachers are weighted by how strongly both quantities move the probability of this single token. This single-token alignment suits mathematics: as shown in Table~\ref{tab:teacher_entropy_stats}, the mathematics teacher has the lowest predictive entropy and a most sharpened token distribution, so its supervision concentrates on the dominant candidate. Code and instruction following involve more positions with several plausible continuations, where alignment on a single token cannot capture how the teacher redistributes probability among competing candidates.

\paragraph{K=32.} Enlarging the support to $k=32$ degrades all three domains. The added candidates carry little probability mass, as the shared top-16 candidates already cover about 99\% of the student's mass according to Table~\ref{tab:novelty_routing_stats}. However, each candidate contributes one coordinate to $\mathbf{e}_{i,t}$ and $\mathbf{d}_{i,t}$ regardless of its probability, and log-ratios amplify differences between small probabilities. For example, a tail token whose probability drops from $10^{-3}$ under the student to $10^{-6}$ under a sharpened teacher yields $|d_{i,t}(v)|\approx6.9$, whereas a head token moving from $0.5$ to $0.4$ yields only $0.22$. Because these probabilities are tiny, even small absolute fluctuations also cause large changes in their log-ratios. As more tail candidates enter the support, the alignment $\langle\mathbf{e}_{i,t},\mathbf{d}_{i,t}\rangle$ is increasingly dominated by near-zero-probability tokens rather than the dominant candidates the student is likely to generate, making both teacher selection and weighting noisy.

\subsection{ExpertAlign Selection Coverage and Training Dynamics}
\label{app:expert_align_skip}

\begin{figure}[h]
    \centering
    \begin{minipage}[t]{0.48\linewidth}
        \centering
        \includegraphics[width=\linewidth]{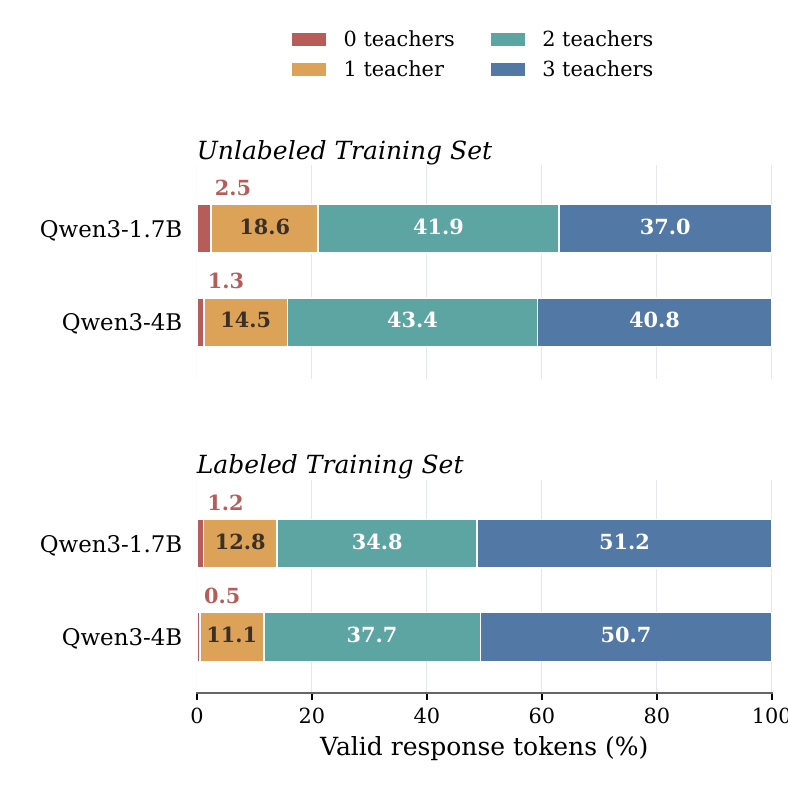}
        \vspace{-1mm}\textbf{(a)} Retained-teacher count distribution
    \end{minipage}\hfill
    \begin{minipage}[t]{0.48\linewidth}
        \centering
        \includegraphics[width=\linewidth]{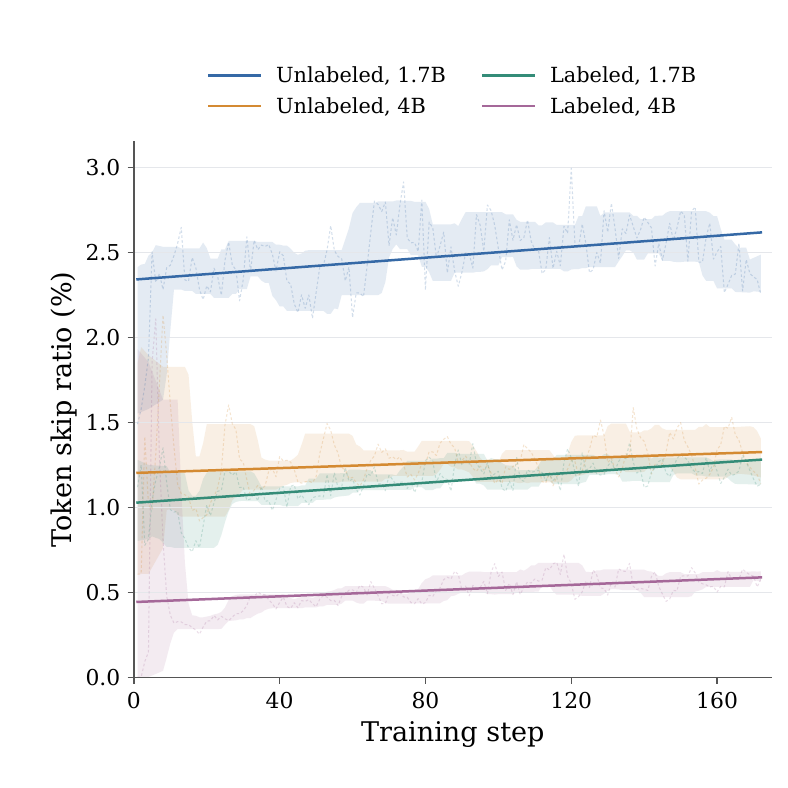}
        \vspace{-1mm}\textbf{(b)} All-teacher skip ratio over training
    \end{minipage}
    \caption{ExpertAlign teacher selection and supervision coverage. (a) Trajectory-mean fraction of valid response tokens with $K_t\in\{0,1,2,3\}$. (b) Evolution of $K_t=0$ over training. Pale dashed curves are logged observations and lines are ordinary-least-squares fits over the displayed range.}
    \label{fig:expert_align_selection_analysis}
\end{figure}

\paragraph{Selective routing with broad supervision coverage.}
Let $K_t=|\mathcal{E}_t|$ denote the number of teachers retained for a valid response token. Figure~\ref{fig:expert_align_selection_analysis}(a) reports the trajectory mean of the token fraction for each value of $K_t$. Across the four settings, 37.02--51.24\% of tokens retain all three teachers; equivalently, ExpertAlign selects a strict subset on 48.76--62.98\% of tokens. Most of these selectively routed tokens remain supervised: 47.61--60.50\% of all tokens retain one or two teachers, whereas only 0.52--2.48\% have $K_t=0$. Consequently, 97.52--99.48\% of valid response tokens receive a signal from at least one teacher. The dominant behavior of the gate is therefore to alter teacher composition, rather than to remove the token from the OPD objective.

\paragraph{Fully filtered tokens increase mildly over training.}
Although $K_t=0$ remains rare throughout training, Figure~\ref{fig:expert_align_selection_analysis}(b) shows a positive ordinary-least-squares slope in each setting. The estimated increases per 100 steps are 0.161 and 0.071 percentage points for the unlabeled 1.7B and 4B students, and 0.148 and 0.084 percentage points for their labeled counterparts. These slopes summarize a mild increasing tendency, and we treat it as a descriptive training dynamic and examine whether it is compatible with the geometry of the ExpertAlign gate below.

\paragraph{Geometric interpretation.}
The observed tendency is compatible with a geometric interpretation of the ExpertAlign gate. Define the student's base-relative displacement on the same support as $\mathbf{q}_t
    =\left[\log p^S_t(v)-\log p^{\mathrm{base}}_t(v)\right]_{v\in S^S_t}$. For a nonzero specialization vector $\mathbf{e}_{i,t}\neq\mathbf{0}$, the student's normalized progress along that direction as $r_{i,t}=\langle\mathbf{e}_{i,t},\mathbf{q}_t\rangle/\lVert\mathbf{e}_{i,t}\rVert_2^2$. Because $\mathbf{d}_{i,t}=\mathbf{e}_{i,t}-\mathbf{q}_t$, the alignment tested by the gate becomes
\begin{equation}
    \langle\mathbf{e}_{i,t},\mathbf{d}_{i,t}\rangle
    =\lVert\mathbf{e}_{i,t}\rVert_2^2
    -\langle\mathbf{e}_{i,t},\mathbf{q}_t\rangle
    =\lVert\mathbf{e}_{i,t}\rVert_2^2(1-r_{i,t}).
    \label{eq:expert_align_progress}
\end{equation}
As the training progresses, the student approaches teacher $i$ and the teaching vector $\mathbf{d}_{i,t}$ contracts, making its inner product with $\mathbf{e}_{i,t}$ more likely to fall below the numerical margin. Besides, when $r_{i,t}\geq 1$, the student's projection reaches or exceeds the teacher's base-relative displacement along this direction, and the alignment also becomes non-positive. Both cases indicate that the student has absorbed the teacher's specialization sufficiently for that teacher to provide little or no positively aligned marginal correction at the current token.

\end{document}